# ASVRI-Legal: Fine-Tuning LLMs with Retrieval-Augmented Generation for Enhanced Legal Regulation

One Octadion[1,*] Bondan Sapta Prakoso[1,2] Nanang Yudi Setiawan[1,2] Novanto Yudistira[3,]

[1] *Jalin Mayantara Indonesia*
[2] *Information System Department, Faculty of Computer Science, Brawijaya University*
[3] *Informatics Engineering Department, Faculty of Computer Science, Brawijaya University*

[*]*Corresponding author. Email:* yudistira@ub.ac.id

**ABSTRACT**

In this study, we explore the fine-tuning of Large Language Models (LLMs) to better support policymakers in their crucial work of understanding, analyzing, and crafting legal regulations. To equip the model with a deep understanding of legal texts, we curated a supervised dataset tailored to the specific needs of the legal domain. Additionally, we integrated the Retrieval-Augmented Generation (RAG) method, enabling the LLM to access and incorporate up-to-date legal knowledge from external sources. This combination of fine-tuning and RAG-based augmentation results in a tool that not only processes legal information but actively assists policymakers in interpreting regulations and drafting new ones that align with current needs. The results demonstrate that this approach can significantly enhance the effectiveness of legal research and regulation development, offering a valuable resource in the ever-evolving field of law.

## 1. INTRODUCTION

The field of artificial intelligence (AI) has seen rapid advancements in recent years, with transformative impacts across various domains, including the legal sector. Since its early applications, such as legal Q&A systems [1], AI has demonstrated its potential to assist legal professionals by providing quick and accurate answers to complex legal inquiries. However, the evolution of AI in the legal field is far from static; it is continuously progressing, moving beyond simple tasks to more complex and nuanced applications [2].

AI's potential in the legal domain is particularly promising as we witness the rise of Large Language Models (LLMs), which have shown significant improvements in their ability to follow intricate instructions and generate human-like text. This evolution opens up new possibilities for AI to play a more integral role in the legal profession, where the demands are often multifaceted and extend far beyond basic question-and-answer tasks. Legal work encompasses a wide range of activities, including the analysis of overlapping legal provisions, the extraction of key legal elements, conducting detailed Q&A, summarizing legal texts and regulations, and drafting new or revised legal articles that can serve as foundations for future regulations.

Traditional AI models struggled with the deep understanding of legal principles, context, and language required for many legal tasks, challenged by legal language's complexity and specific reasoning demands. However, with advancements in LLMs [3] [4] [5], we now have tools to effectively understand and process this complex language, primarily assisting professionals by automating tasks and providing data-driven insights for more efficient navigation of complex work.

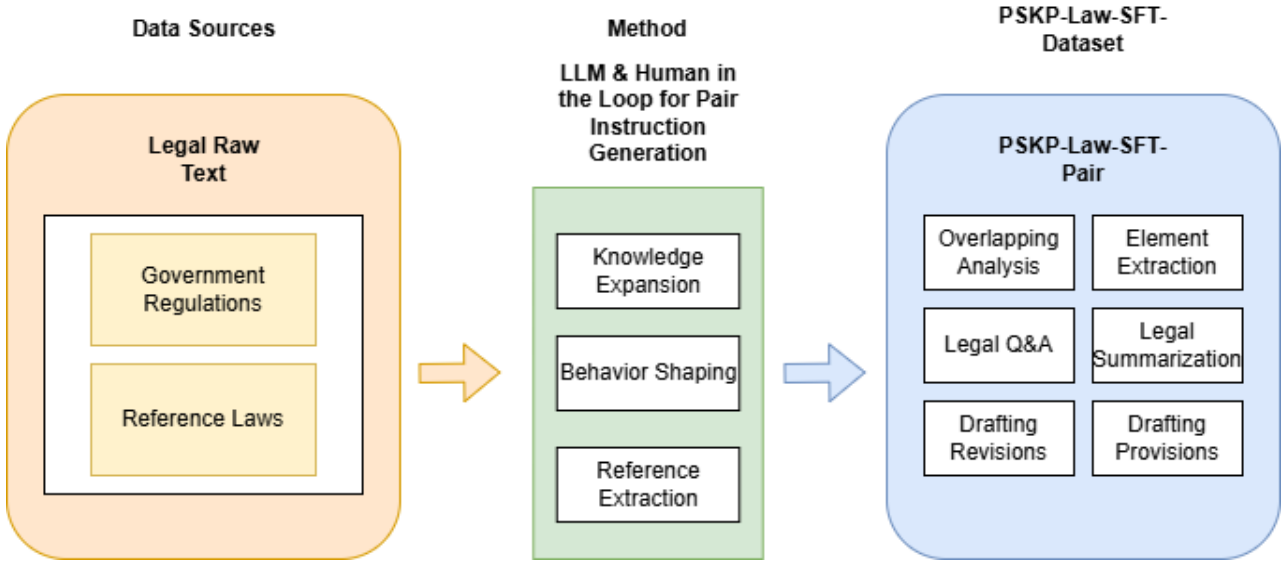

*Figure 1: Construction of Dataset*

experts to focus on higher-level decision-making and creative problem-solving.

Recognizing the potential of LLMs in this domain, we embarked on a project to fine-tune an LLM specifically for the legal and regulatory sector. Our goal was to enhance the model's reasoning capabilities within this specialized domain by using a supervised dataset tailored to the unique tasks required in legal work. This fine-tuning process is designed to improve the model's ability to interpret, analyze, and generate legal texts, making it a more effective tool for legal professionals [6].

In addition to fine-tuning, we incorporated the Retrieval-Augmented Generation (RAG) method into the model. The RAG method allows the LLM to access and integrate external knowledge, ensuring that it remains informed with the most current legal information [7]. This is particularly important in the legal field, where regulations and legal precedents are constantly evolving. By combining domain-specific fine-tuning with the ability to retrieve and incorporate external knowledge, we aim to create a tool that not only understands legal texts but also provides accurate and up-to-date information, reducing the likelihood of hallucinations (i.e., generating incorrect or nonsensical information).

This LLM, with its enhanced understanding of legal and regulatory domains and its ability to access external knowledge, has the potential to significantly improve the efficiency and effectiveness of legal professionals. It offers a reliable resource for those involved in legal analysis, regulation drafting, and policy-making, providing assistance in tasks that require a deep understanding of legal language and reasoning.

In this study, we present our approach to fine-tuning an LLM for the legal domain, detailing the methodologies used and the outcomes achieved. We believe that this work represents a significant step-forward in the application of AI in the legal field, offering new tools and capabilities for legal professionals to leverage in their work.

## 2. RELATED WORKS

Large Language Models (LLMs) have achieved high performance in linguistic tasks and reasoning. However, general LLMs often fall short in terms of depth of knowledge and accuracy. They also tend to be resource-intensive, which impacts their efficiency. In contrast, domain-specific LLMs can offer greater relevance by focusing on specialized training data, but they may encounter issues such as overfitting. Consequently, enhancing LLMs for specific domains, like the legal field, presents its own set of challenges and opportunities [8].

Recent advancements in legal domain LLMs show significant progress. LawGPT is an open-source model tailored for Chinese legal applications, addressing the limitations of general LLMs by using legal-oriented pre-training and supervised fine-tuning on specialized datasets. This approach improves performance over models like LLaMA 7B in legal tasks [9].

Similarly, Lawyer LLaMA adapts general LLMs for legal use by incorporating domain knowledge during continual training and using expert-designed supervised fine-tuning tasks. It also includes a retrieval module to reduce hallucinations by extracting relevant legal articles. This model demonstrates that expert-written data is more effective than data generated by models like ChatGPT, highlighting the importance of domain-specific enhancements [10] [11].

Chatlaw introduces a Mixture-of-Experts (MoE) model and a multi-agent system to improve AI-driven legal services. By integrating knowledge graphs and Standardized Operating Procedures (SOPs), Chatlaw enhances accuracy and reduces errors and hallucinations, outperforming GPT-4 in legal evaluations and consultations [12].

SaulLM-7B is another notable development, being the first LLM explicitly designed for legal text comprehension and generation. Built on the Mistral 7B architecture and trained on a large legal corpus, SaulLM-7B demonstrates state-of-the-art proficiency in processing legal documents. A novel instructional fine-tuning method further enhances its performance in legal tasks, showcasing its potential as a robust tool in the legal domain [13].

## 3. DATASETS

To develop a robust legal Large Language Model (LLM), we created a specialized supervised fine-tuning dataset tailored to the legal domain. This dataset is designed to enhance the model's ability to both comprehend and generate content related to legal regulations and statutes. Unlike typical datasets used for general Q&A models, our dataset addresses the unique challenges posed by legal language and reasoning.

The dataset, whose construction process is illustrated in Figure 1, is organized into instruction-output pairs corresponding to various legal tasks. These tasks are carefully selected to cover the broad spectrum of capabilities required for effective legal analysis and document generation. For example, the dataset includes tasks focused on analyzing overlapping legal provisions, extracting key elements from legal texts, and handling legal Q&A with a nuanced understanding of context and terminology. Additionally, the dataset incorporates tasks for summarizing legal provisions and generating suggestions for revising existing laws or drafting new ones, as detailed in Table 1.

By training the model on this diverse set of tasks, we aim to equip it with a comprehensive understanding of the legal domain. The dataset not only improves the model's fundamental comprehension of legal language but also enhances its ability to generate accurate and contextually appropriate legal content. This dual focus on understanding and generation is crucial for creating a legal LLM that can effectively assist legal professionals in a variety of complex tasks.

| Dataset | Task | Size | Scenario |
| --- | --- | --- | --- |
| ASVRI-PSKP-Legal-SFT | Legal overlapping analysis | 1087 | Identifying conflicts and redundancies in legal texts |
| | Element extraction | 890 | Extracting key legal elements (e.g., clauses, parties, dates) from documents |
| | Legal Q&A | 1794 | Answering legal queries based on case law or statutes |
| | Legal summarization | 1415 | Summarizing lengthy legal documents into concise briefs |
| | Drafting revisions | 1571 | Revising and improving existing legal drafts |
| | Drafting provisions | 1750 | Creating new legal provisions or clauses |
| **Total** | | 8507 | |

*Table 1: Datasets overview*

## 3.1. Data Sources

Our dataset is primarily derived from government regulations and laws, specifically focusing on the domain managed by the Pusat Standar dan Kebijakan Pendidikan (PSKP), which oversees educational standards and policies.

1. **Government Regulations on PSKP Policies**

   The first data source consists of government regulations within the PSKP domain. These regulations encompass various legal provisions that govern educational standards and policies, including mandates like Compulsory Education, Education Funding, and related topics. The dataset was carefully curated to capture the breadth and depth of these regulations, ensuring that the model could understand and generate content aligned with current educational policies.

2. **Reference Laws**

   In addition to government regulations specific to PSKP, our dataset also includes legal provisions from reference laws that serve as the foundation for these regulations. This integration ensures that each regulation is grounded in its legal basis, allowing the model to explore and understand the relationships between different legal provisions. By incorporating these reference laws, we aimed to create a more comprehensive and interconnected dataset, enriching the model's ability to navigate and generate content that respects the hierarchical nature of legal texts.

## 3.2. Instruction Dataset Generation

To effectively fine-tune our Large Language Model (LLM) for the legal domain, we created a dataset in the form of instruction pairs, which were specifically designed for supervised fine-tuning. This process involved transforming raw legal texts into instruction-output pairs tailored to the tasks we aimed to train the model on [14].

We utilized OpenAI's GPT-3.5-turbo model to assist in this transformation process. The legal texts, which consisted of various regulations and laws, were systematically converted into structured instruction pairs. Before these legal texts were used by OpenAI's GPT-3.5-turbo model, we applied several methods to process and structure the data effectively :

1. **Chunking Method**

   We implemented a chunking technique where the legal texts were divided into chunks of predefined sizes. Each chunk was then processed individually by the GPT-3.5-turbo model. This method allowed us to generate diverse data representations, as the content within each chunk varied depending on its specific segment [15].

2. **Hierarchical Data Segmentation**

   To enhance the structure and coherence of the dataset, we segmented the legal texts into different hierarchical levels: per article, per clause, and per chapter [16] This approach provided more control over the data being fed into the model, enabling a clearer representation of relationships between various legal elements. By structuring the data in this way, we ensured that the resulting dataset was not only more organized but also capable of capturing and reflecting the intricate connections between different legal provisions.

Once the data was processed and segmented, we used GPT-3.5-turbo with specifically crafted prompts to convert the segments into instruction pairs as shown in Table 2. Each task, such as overlapping analysis, element extraction, Q&A, summarization, drafting revisions, and creating new legal provisions, had its own tailored prompt [17]. The outcome was a comprehensive set of input-output pairs designed to fine-tune the LLM effectively for legal tasks.

Input:

1. Subject [or] Create 3 data based on the **context** above with the input and output format focused on the task of element extraction from an article. Include references to the government regulation (PP), article, and clause. The input should be as follows:

Object [or] Action [or] Intention [or] Sanction [or] Time [or] Duration [or] Location [or] Procedure of the article in {insert the full text of the article and clause from the data}.

Then, adjust the output to match the input (if the input is "Subject," then the output should be "Subject"), with the content being the elements contained in the article as follows:

Output:

Subject of the Article: {Subject of the Article}\n [or] Object of the Article: {Object of the Article}\n [or] Action: {Action}\n [or] Intention: {Intention}\n [or] Sanction: {Sanction}\n [or] Time and Duration: {Time and Duration}\n [or] Location or Place: {Location or Place}\n [or] Procedure: {Procedure}\n

For each data entry, create it as a JSON object in the following format:

{

  'input': the question,

  'output': the answer,

}

| *No* | *Question* | *Answer* |
|---|---|---|
| **1.** | What were the important problems and challenges faced in the 2015-2019 period? | Important problems and challenges faced in the 2015-2019 period include the implementation of quality 12-year compulsory education, improving the quality of learning, enhancing teacher management and teacher education, increasing access to and quality of early childhood education, improving work skills and adult education, and strengthening education governance and efficiency in education financing. |
| **2.** | What is stipulated regarding the study load to become a teacher in a Kindergarten (TK), Raudhatul Athfal (RA), Special Needs Kindergarten (TKLB), or other equivalent educational units for those with an S-1 or D-IV education background? | The study load to become a Teacher in a Kindergarten (TK), Raudhatul Athfal (RA), Special Needs Kindergarten (TKLB), or other equivalent educational units for those with an S-1 or D-IV education background is 18 (eighteen) to 20 (twenty) semester credit units. |
| **3.** | Based on consideration b, what article revision suggestion can further regulate the bodies involved in the development of the National Education Standards? | The recommended article for revision is Article 56 concerning Closing Provisions, to further regulate the bodies involved in the development of the National Education Standards. |

*Table 2: Specific task prompt example*

## 4. ASVRI LEGAL LLM

To create a specialized model of ASVRI (a personal virtual assistant) for understanding PSKP regulations, we fine-tuned a Large Language Model (LLM) using supervised learning techniques. Additionally, we incorporated Retrieval-Augmented Generation (RAG) to enable the model to access and utilize relevant legal documents. This approach enhances the model's ability to interpret and generate accurate legal content by combining domain-specific training with effective retrieval methods.

### 4.1. Supervised Fine-Tuning

We developed the ASVRI Legal LLM based on the open-source LLaMA2-7B and WizardLM-13B models, each with approximately 7 and 13 billion parameters. To enhance the model's understanding of legal regulations and laws, we conducted supervised fine-tuning using a carefully constructed dataset [18].

For effective fine-tuning, we employed the PEFT (Parameter Efficient Fine-Tuning) method, specifically LoRA (Low-Rank Adaptation), allowing us to adjust only a portion of the model's weights. This approach helps minimize catastrophic forgetting (the tendency of AI models to lose previously learned information when learning new tasks) and reduces computational requirements [19].

The fine-tuning process involved using a per-device batch size of 2 for LLaMA2-7B and 1 for WizardLM-13B, and training was conducted for 3 epochs. The maximum source length was set to 2048 tokens, and the maximum target length to 1024 tokens. Training was performed on a 1xT4 and 4xT4 GPUs.

### 4.2. Retrieval Augmented Generation

To build an effective AI system that assists users in resolving legal matters, it is essential for the model to have access to all relevant regulations and laws. This is particularly important since the legal knowledge used during the model's training might become outdated or miss recent amendments. By implementing the Retrieval Augmented Generation (RAG) method, we can help the model stay current and reduce the risk of hallucinations [20] [21]

As seen in Figure 2, we began by creating a knowledge base comprising regulations and laws specifically related to education, including government regulations and laws. Each of these documents was converted into vectors and stored in the knowledge base. When a user inputs a query, the retriever searches this knowledge base and returns the top-k most relevant documents based on similarity scores.

These retrieved documents are then fed into our ASVRI Legal LLM, alongside the user's input and a system prompt. By referencing this updated and relevant knowledge base, the model can better understand the context and provide more accurate and aligned responses [15].

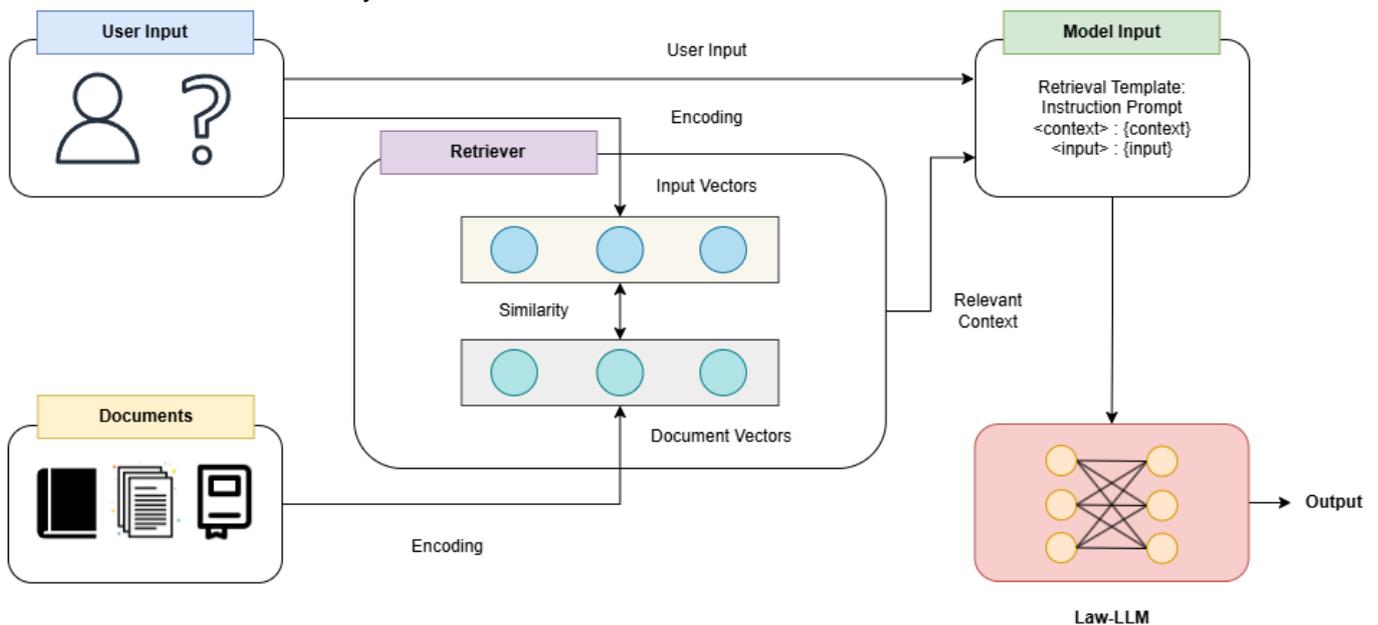

*Figure 2: Overview of Retrieval Augmented Generation Law LLM*

## 5. EVALUATIONS

Evaluating the performance of our ASVRI Legal LLM is crucial to understanding its capability in handling complex legal tasks [22], particularly in the domain of regulations and laws related to PSKP.

### *5.1. Quantitative Evaluation*

For the quantitative evaluation, we employed standard natural language processing metrics, specifically BLEU and METEOR scores. These metrics were applied to a test dataset that consisted of 20% of the data originally used for training. BLEU score was utilized to measure the precision of n-grams or sequences of N words in the model's output compared to the reference text, providing an indicator of the model's ability to generate text that closely matches the expected legal language. Similarly, the METEOR score was used to evaluate both precision and recall, considering synonyms and stemming, which is particularly useful for legal text where exact word matches might not always reflect the model's understanding.

### *5.2. Qualitative Evaluation*

Beyond the quantitative metrics, we conducted a qualitative evaluation by manually analyzing the model's responses. This process involved reviewing the model's outputs to assess their contextual appropriateness, legal relevance, and overall quality. Our manual review focused on how well the model could understand and apply legal knowledge to generate coherent and accurate legal texts, particularly in more nuanced scenarios where legal interpretation and judgment are required.

By evaluating the model's responses in real-world legal tasks, we were able to gain insights into its practical utility and identify areas where it excels or needs improvement. This subjective analysis was crucial in ensuring that the model not only meets technical benchmarks but also performs well in practical applications, providing reliable assistance in legal matters.

## 6. EXPERIMENTS

To demonstrate the capabilities of our trained ASVRI Legal LLM, we conducted a series of experiments using the evaluation dataset described earlier. These experiments compared the performance of our ASVRI Legal LLM with several benchmark models to assess its effectiveness in understanding and generating legal text, particularly in the PSKP domain.

We tested the following models: ASVRI Legal LLM based on LLaMA 2 with 7 billion parameters, ASVRI Legal LLM based on WizardLM with 13 billion parameters, the base LLaMA 2 model with 7 billion parameters without any legal-specific fine-tuning, OpenAI's GPT-3.5 Turbo, and OpenAI's GPT-4.

Each model processed the same legal dataset, and their outputs were evaluated using both objective and subjective metrics, as discussed in the previous section. This comparison highlights the strengths and weaknesses of our fine-tuned ASVRI Legal LLM in relation to other well-established models, providing a comprehensive understanding of its performance in legal contexts.

### *6.1. Results*

The experiment results, as comprehensively detailed in Table 3, serve to demonstrate the varying degrees of effectiveness exhibited by the different models when tasked with generating legal text specifically within the PSKP domain. Commencing with the quantitative evaluation, the ASVRI-Legal-Llama 2 model, equipped with 7 billion parameters and employing the Retrieval-Augmented Generation (RAG) method without specific fine-tuning on our legal dataset, initially achieved a BLEU score of 0.01 and a Meteor score of 0.09.

These baseline scores, while modest, indicate a foundational, albeit basic, level of understanding of legal text structure and vocabulary. It is crucial to note that these scores were rigorously derived from a 20% test dataset that was meticulously held out during the training phase, a standard practice to ensure an objective and unbiased assessment of the model's ability to generalize to unseen data, thereby preventing overfitting.

Subsequently, the model's performance metrics indicated a significant improvement upon the introduction of fine-tuning with our specialized legal dataset. This targeted training elevated its BLEU score to 0.07 and its Meteor score to 0.24. Furthermore, when this fine-tuning was synergistically combined with the RAG method, the model's capabilities were further enhanced, reaching a BLEU score of 0.13 and a Meteor score of 0.34. This notable progression in scores reflects the model's improved precision in generating n-grams that match reference texts and enhanced recall in capturing relevant semantic meaning, crucial for the nuanced demands of legal text generation.

| Model | Size | Method | Metrics | |
|---|---|---|---|---|
| | | | BLEU | Meteor |
| **ASVRI-Legal-Llama 2** | 7B | RAG | 0,01 | 0,09 |
| **ASVRI-Legal-Llama 2** | 7B | Fine-tune | 0,07 | 0,24 |
| **ASVRI-Legal -Llama 2** | 7B | Fine-tune + RAG | 0,13 | 0,34 |
| **ASVRI-Legal -Wizardlm** | 13B | Fine-tune + RAG | 0,15 | 0,37 |
| **GPT-3.5-Turbo** | 175B | RAG | 0,24 | 0,4 |
| **GPT-3.5-Turbo** | 175B | Fine-tune + RAG | 0,25 | 0,48 |
| **GPT-4** | - | RAG | 0,17 | 0,46 |

*Table 3: Results compared with general and ASVRI legal LLMs*

The Legal-WizardLM model, with 13 billion parameters, performed better across both metrics, achieving a BLEU score of 0.15 and a Meteor score of 0.37 when using fine-tuning and RAG, indicating that the larger model size contributes to a better understanding and generation of legal language. The GPT-3.5 Turbo model outperformed the fine-tuned ASVRI Legal LLMs, achieving a BLEU score of 0.24 and a Meteor score of 0.40 using RAG, and these scores improved to 0.25 and 0.48 after fine-tuning. Lastly, GPT-4 showed strong performance with a BLEU score of 0.17 and a Meteor score of 0.46, which, although slightly lower in BLEU, maintained high recall and precision as indicated by the Meteor score.

For the qualitative evaluation, we conducted a manual review of the model outputs to assess their contextual appropriateness and legal coherence. The fine-tuned ASVRI Legal LLMs, particularly ASVRI-Legal-Llama 2 and ASVRI-Legal-WizardLM, demonstrated significant improvements in generating text that aligns with legal standards and terminology, especially in complex scenarios where nuanced legal interpretation is required. However, despite these improvements, these models still fall short when compared to the larger, more generalized models like GPT-3.5 Turbo and GPT-4. These larger models not only excel in quantitative metrics but also deliver superior performance in the structure and content of legal text, making them more reliable for practical legal applications.

## 7. DISCUSSION

The current ASVRI Legal LLM exhibits several limitations, particularly in complex legal tasks such as analyzing overlaps between legal provisions, drafting revisions, and creating new clauses. The model's performance, as indicated by the BLEU and Meteor scores, suggests it is not yet adept at handling nuanced legal content. This indicates a need for further refinement in both model architecture and training to address these complex requirements.

The evaluation metrics used, BLEU and Meteor, while standard, may not fully capture the intricacies of legal text generation. These metrics focus on n-gram overlap and semantic similarity, which may not adequately reflect the model's understanding of complex legal concepts. Future evaluations might benefit from incorporating metrics specifically designed for legal text, which could offer a more precise measure of the model's capabilities.

Practically, the model shows promise but also has limitations. While it can identify conflicts and redundancies in legal texts to some extent, it may require additional tools for enhanced accuracy. It performs reasonably well in extracting key legal elements and answering legal queries, though it may struggle with less common or nuanced questions. Summarizing lengthy documents into concise briefs is feasible, yet the summaries might need refinement for legal precision. For revising drafts and creating new legal provisions, the model provides a useful starting point but often requires human oversight to ensure legal compliance. Future work should focus on enhancing the model's ability to handle these complex tasks and exploring new evaluation metrics tailored to legal contexts.

| Model | Question | Response |
|---|---|---|
| **ASVRI-Legal-Wizardlm** | How are the rights and obligations of the community related to the implementation of the compulsory education program carried out? | 1. The community has the right to: participate in the implementation, supervision, and evaluation of the compulsory education program; and obtain data and information regarding the compulsory education program.<br><br>2. The community is obliged to support the implementation of the compulsory education program.<br><br>3. The rights and obligations of the community as referred to in paragraphs (1) and (2) are carried out in accordance with statutory regulations. |
| | What is the definition of basic education according to Government Regulation? | 1. Basic education is the level of education that underlies secondary education.<br><br>2. Basic education takes the form of Elementary Schools (SD) and Islamic Elementary Schools (MI) or other equivalent forms, as well as Junior High Schools (SMP) and Islamic Junior High Schools (MTs), or other equivalent forms.<br><br>3. Further provisions regarding basic education as referred to in paragraphs (1) and (2) are regulated by Government Regulation. |
| | What must educational units do for prospective students and students? | 1. Educational units are obliged to provide places for prospective Indonesian citizen students who have adequate academic potential and are economically disadvantaged, at least 20% of the total number of new students.<br><br>2. Educational units are obliged to provide scholarships for students with citizenship who have achievements.<br><br>3. The decision on the admission of prospective students to become students is made independently by a meeting of the teachers' council led by the head of the educational unit.<br><br>4. The selection for new student admissions in grade 10 at secondary education units is based on the results of the National Examination, except for students as referred to in Article 81 paragraphs (2), (4), and (5).<br><br>5. In addition to fulfilling the provisions as referred to in paragraph (4), educational units may conduct scholastic aptitude tests for the selection of new student admissions in grade 10.<br><br>6. New student admissions can be carried out every semester for educational units that implement a semester credit system. |

*Table 4: Example of results from finetuned LLM*

## 8. CONCLUSION

This study has explored the fine-tuning of Large Language Models (LLMs) to enhance their utility in the legal and regulatory domain, particularly in supporting policymakers. By leveraging a combination of supervised fine-tuning with domain-specific datasets and the Retrieval-Augmented Generation (RAG) method, we aimed to create a model capable of effectively processing, analyzing, and generating legal texts.

The results from our experiments indicate that while fine-tuning LLMs and integrating RAG significantly improve the model's performance in legal text generation, challenges remain. The ASVRI-Legal-Llama 2 and ASVRI-Legal-WizardLM models, despite showing substantial improvements over their baseline versions, still fall short when compared to larger, more generalized models like GPT-3.5 Turbo and GPT-4. These larger models consistently outperformed the fine-tuned legal-specific models, both in quantitative metrics such as BLEU and Meteor scores and in qualitative assessments of legal coherence and contextual appropriateness.

However, the fine-tuned ASVRI Legal LLMs demonstrated a notable capacity to generate legally relevant text and support legal professionals in tasks such as regulation drafting and legal analysis. This suggests that with further refinement, particularly in handling complex legal scenarios and improving evaluation metrics tailored to legal contexts, these models could become invaluable tools in the legal domain.

In conclusion, while the current fine-tuned models show promise, they require further development to meet the high standards required for practical legal applications. Future research should focus on enhancing model architectures, expanding domain-specific training data, and developing more sophisticated evaluation metrics to better capture the intricacies of legal text generation. The integration of these improvements has the potential to create powerful AI-driven tools that can significantly augment the work of legal professionals and policymakers.

## AUTHORS' CONTRIBUTIONS

One Octadion contributed to conducting the research and writing the initial draft. Bondan Sapta Parkoso took charge of data analysis and research, guiding and providing suggestions for the study, as well as making critical revisions. Nanang Yudi Setiawan was responsible for data analysis and research as well as critical revisions. Novanto Yudistira contributed to data analysis, guided the research process, suggested research concepts, and evaluated the study, among other contributions.

## ACKNOWLEDGMENTS

The authors would like to thank the Jalin Mayantara Indonesia team for their assistance in the research process, both in operational support and in providing the necessary facilities and resources.